\documentclass{article}

   \PassOptionsToPackage{numbers, compress}{natbib}

\usepackage[preprint]{neurips_2022}

\usepackage[utf8]{inputenc} 
\usepackage[T1]{fontenc}    
\usepackage{hyperref}       
\usepackage{url}            
\usepackage{booktabs}       
\usepackage{amsfonts}       
\usepackage{nicefrac}       
\usepackage{microtype}      
\usepackage{xcolor}         

\usepackage{amsmath}
\usepackage{graphicx}
\usepackage{algorithm}
\usepackage{algorithmic}
\usepackage{multirow}

\title{Sparse Adversarial Attack in Multi-agent Reinforcement Learning}

\author{%
  Yizheng Hu \\
  Academy for Advanced Interdisciplinary Studies\\
  Peking University\\
  No.5 Yiheyuan Road, Haidian District, Beijing 100871 \\
  \texttt{1801111560@pku.edu.cn} \\
  \And
  Zhihua Zhang \\
  School of Mathematical Sciences \\
  Peking University \\
  No.5 Yiheyuan Road, Haidian District, Beijing 100871 \\
  \texttt{zhzhang@math.pku.edu.cn} \\
}

\begin{document}

\maketitle

\begin{abstract}
  Cooperative multi-agent reinforcement learning (cMARL) has many real applications, but the policy trained by existing cMARL algorithms is not robust enough when deployed. There exist also many methods about adversarial attacks on the RL system, which implies that the RL system can suffer from adversarial attacks, but most of them focused on single agent RL. In this paper, we propose a \textit{sparse adversarial attack} on cMARL systems. We use (MA)RL with regularization to train the attack policy. Our experiments show that the policy trained by the current cMARL algorithm can obtain poor performance when only one or a few agents in the team (e.g., 1 of 8 or 5 of 25) were attacked at a few timesteps (e.g., attack 3 of total 40 timesteps).
\end{abstract}

\section{Introduction}
\label{Introduction}

Many real-world sequential decision problems involve multiple agents in the same environment, where all agents work together to accomplish a certain goal. Cooperative multi-agent reinforcement learning (cMARL) is a key technique for solving these problems. In recent years, cMARL has been applied to many fields, such as traffic light control \citep{wang2021adaptive}, autonomous driving \citep{shalev2016safe}, wireless networking \citep{nasir2019multi}, etc. 

However, many prior works about adversarial attacks on single agent RL have already shown that the policies trained by most current single agent RL algorithms are not robust enough, thus one cannot use them with confidence in real-world applications. But to the best of our knowledge, there are very few prior works on attacking MARL policies. \citet{lin2020robustness} and \citet{guo2022towards} showed that a MARL policy can perform awful when one agent behaved adversarially against other agents.

Unlike many prior works which attack RL agents at every timesteps, in this paper we attempt to look further. More specifically, we want to find whether the policy trained by the current MARL algorithm like QMIX can be significantly affected, when only a few agents in the team were attacked at a few timesteps during the whole episode, i.e., \textit{sparse attack}. Here ``sparse'' has two meanings: sparseness in agent dimension and sparseness in time dimension.

This result is important for judging whether the current MARL policy can be deployed into real-world scenarios, because in many existing cMARL algorithms the team can only guarantee to get a high reward when all agents execute their optimal policy accurately. But this may not be true for real-world scenarios. For a machine team, the machines may malfunction sometimes; for a human team, the teammates may not be absolutely credible, because the competitors may plant a spy who will sabotage the team.

But of course, if a machine fails frequently, it will be replaced by a better machine soon; and if a spy sabotages the team frequently, he/she will be easy to be spotted and kicked out from the team. Real-world machines may more likely to fail only occasionally, and a spy can only sabotage occasionally to avoid being kicked out. This is why we focus on sparse attack. 

Prior sparse attack work mainly focused on single agent RL. Most of them used some rule-based method to get the timestep of the attack, which can be shown to be sub-optimal (see \ref{sec4.1}). In this work, we attempt to use (MA)RL algorithms to learn the optimal sparse attack policy. To the best of our knowledge, our work is the first work about optimal sparse attack in the MARL environment.

Our contributions are summarized as follows:
\begin{itemize}
	\item We perform adversarial attacks in the MARL environment, and consider both single agent attack and multi agent attack. 
	\item We propose a (MA)RL based optimal sparse attack method in the MARL environment, which outperforms existing rule based sparse attack methods.
	\item Our experiments show that the current MARL algorithms, at least for QMIX, are not robust enough.
\end{itemize} 

\section{Related Work}
\label{Related Work}

\paragraph{MARL algorithms}

Existing cMARL algorithms can be roughly divided into two types: policy-based and value based. Representative policy-based algorithms include COMA \cite{foerster2018counterfactual}, MADDPG \cite{lowe2017multi}, etc; and representative value-based methods include VDN \cite{sunehag2017value} QMIX \cite{rashid2018qmix}, QTRAN \cite{son2019qtran}, ROMA \cite{wang2020roma}, Weighted-QMIX \cite{rashid2020weighted}, etc.

\paragraph{Adversarial attacks in RL}


Adversarial attacks in RL can also be roughly divided into two types: attack agent (e.g., perturb agent's observation or action) or modify environment. \textbf{1)} For attacking agent, early works like \cite{behzadan2017vulnerability,huang2017adversarial} used simple adversarial example algorithm to attack agent's state. Later works like \cite{tretschk2018sequential,russo2019optimal} used RL to learn the perturbation of agent's state that can minimize agent's reward. \citet{lee2019spatiotemporally} attacked agent's action directly. \citet{xiao2019characterizing,lin2017tactics,hussenot2019targeted} tried different adversarial attack task (same perturb across time; mislead agent to certain state). \textbf{2)} For modifying environment, \citet{mankowitz2019robust,mankowitz2020robust,abdullah2019wasserstein,zhang2020robustMARL,yu2021robust} provided different kinds of environment modification.

\paragraph{Sparse attack in RL}

\citet{lin2017tactics,qu2019minimalistic,yang2020enhanced} used rule based methods to decide attack steps (based on the ``difference'' or ``entropy'' of the agent's policy or value function). \citet{behzadan2019adversarial,qiaoben2021strategically} used RL to learn the attack steps and then applied adversarial example at those steps. \citet{sun2020stealthy} used RL to learn both the attack step and the perturbation. All these works are conducted in single agent RL.

\paragraph{Adversarial attacks and adversarial training in MARL}

\citet{lin2020robustness,guo2022towards} attacked one agent in the cMARL environment. \citet{pham2022evaluating} used model based method to attack cMARL agents. \citet{li2019robust,sun2021romax,nisioti2021robust} did mini-max adversarial training in the cMARL environment, which assumed some agents may behave adversarially against other agents. All these works assumed agent(s) can behave adversarially at any timestep.

\section{Problem Formulation}
\label{Problem Formulation Section}


A standard cooperative MARL system contains $n$ agents that can be described as a multi-agent Markov decision processes (multi-agent MDP, MMDP) $(\mathcal{S},\{\mathcal{A}_i\}_{i=1}^n,\{\mathcal{O}_i\}_{i=1}^n,R,P)$. Here $\mathcal{S}$ is the environment's global state space. Each agent $i$ has its own action space $\mathcal{A}_i$ and observation space $\mathcal{O}_i$. At each time step $t$, agent $i$ first observes $o_{i,t}\in\mathcal{O}_i$ based on the current global state $s_t\in\mathcal{S}$. The observation $o_{i,t}$ can be the same as $s_t$ if the environment is fully observable, or be different if the environment is partially observable. After that, agent performs action $a_{i,t}$ based on its policy $\pi_i(\cdot|h_{i,t})$, $h_{i,t}$ is agent $i$'s all historical information until $t$. The team receives reward $r_t=R(s_t,a_{1,t}, \ldots,a_{n,t})$, and then the environment transfers to next state $s_{t+1}\sim P(\cdot|s_t,a_{1,t}, \ldots, a_{n,t})$. The team's goal is to maximize its expected total reward $\mathbb{E}_{s_0\sim p_0,\mathbf{a}_{t}\sim\mathbf{\pi},s_{t+1}\sim P}\left[\sum_t R(s_t,\mathbf{a}_{t})\right]$. 

To simplify the notation, we may use $\mathbf{a}_t$ to refer to $a_{1,t}, \ldots, a_{n,t}$, use $\mathbf{a}_{\{i,j,k\},t}$ to refer to agent $\{i,j,k\}$ 's action, and use $\mathbf{a}_{-\{i,j,k\},t}$ to refer all except agent $\{i,j,k\}$ 's action.

For partially observable environments, we may assume the global state $s_t$ is known during MARL's training period, but not during its testing period. The MARL algorithm can use global state information to help agents learn its policy, but each agent must be able to perform its policy only with its own observation. This is a common setting in many existing cMARL algorithms.


If the attacker decide to attack $m\in[1,n]$ agent(s) in the team (denoted by $\mathbf{k}=\{k_1,\cdots,k_m\}$, $k_i\in[1,n]$ for $i\in[1,m]$), the sparse adversarial attack task can be formulated as a constrained optimization problem:

  \begin{equation}
    \begin{aligned}
      \min_{\mathbf{a}_{\mathbf{k},t}} &\mathbb{E}_{s_0\sim p_0,s_{t+1}\sim P}\left[\sum_t R(s_t,\mathbf{a}_{\mathbf{k},t},\mathbf{a}_{-\mathbf{k},t}^*)\right] \\
      \text{s.t. } &\sum_{i=1}^m\sum_t 1_{[a_{k_i,t}\ne a_{k_i,t}^*]} \leq N.
    \end{aligned}
  \end{equation}

where $a^*$ is the optimal policy without attack, and $N$ is the maximum number of attack steps. This optimization problem is a constrained (MA)RL problem. 

Moreover, since our goal is to find out ``whether MARL policy can be significantly affected by a little modification'', the maximum number of attack steps, i.e., $N$, is also a hyper-parameter that needs to be tuned. Instead of solving the constrained (MA)RL and tune the hyper-parameter $N$, we can just relax the hard constraint $\sum_{i=1}^m\sum_t 1_{[a_{k_i,t}\ne a_{k_i,t}^*]} \leq N$ into a soft regularization: 

\begin{equation}
\label{eq:reg}
  \min_{\mathbf{a}_{\mathbf{k},t}} \mathbb{E}_{s_0\sim p_0,s_{t+1}\sim P}\left[\sum_t R(s_t,\mathbf{a}_{\mathbf{k},t},\mathbf{a}_{-\mathbf{k},t}^*)+\lambda \sum_{i=1}^m\sum_t 1_{[a_{k_i,t}\ne a_{k_i,t}^*]}\right].
\end{equation}

The soft regularized problem is enough for our goal. In the regularized problem, $\lambda$ can be used to control the sparsity of attack indirectly, large $\lambda$ leading to less attack.

Most existing works on adversarial attack on RL policy were targeted on attacking state or observation. Unlike this, we make a stronger assumption: the attacker can directly modify agent's action. This assumption is reasonable under sparsity regularization, and it does also reflects some real-world scenarios. A faulty machine may perform sub-optimal actions even when the observation is correct, and a spy may perform any action that can sabotage the team, as long as he/she doesn't do this frequently. 

\section{Methodology}
\label{sec4}

\subsection{Rule based sparse attack is sub-optimal}
\label{sec4.1}

Some prior work about sparse attack in RL used rule based method to choose the attack step. For example, one calculates $\delta_t=\max_a \pi(a|s_t)-\min_a \pi(a|s_t)$ \citep{lin2017tactics} or $\delta_t=\sum_{i=0}^{m} \frac{p\left(a_{t}^{i}\right) \cdot \log p\left(a_{t}^{i}\right)}{\log m}$ \citep{qu2019minimalistic} first and then attacks $t$ with high $\delta_t$. For Q learning, both the methods used $\text{softmax}(Q)$ to calculate $\delta_t$. These two methods have a certain intuitive meaning: higher $\delta_t$ means policy has higher confidence in its optimal action, which potentially means step $t$ is more important and more worthwhile to be attacked. But both the methods are sub-optimal, at least for attacking Q learning agent. Here's an example:

\paragraph{Example 1}

Consider a $\{0,1\}$ action game with total $T$ timesteps. At timestep $t$ there are $2^t$ states. The state transit function is deterministic and different action sequences will lead to different states. The game can be described by a binary tree. The agent only receives reward at the last step, with no discount factor.

\begin{figure}
	\centering
  \begin{minipage}[t]{0.48\textwidth}
    \centering
    \includegraphics[width=0.9\columnwidth]{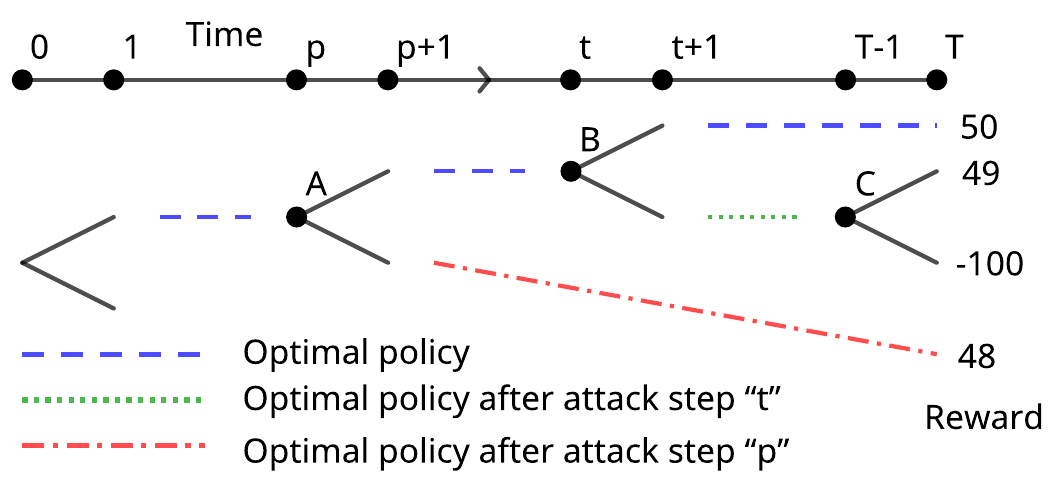}
		\caption{Example 1}
		\label{example1}
    \end{minipage}
    \begin{minipage}[t]{0.48\textwidth}
    \centering
    \includegraphics[width=0.9\columnwidth]{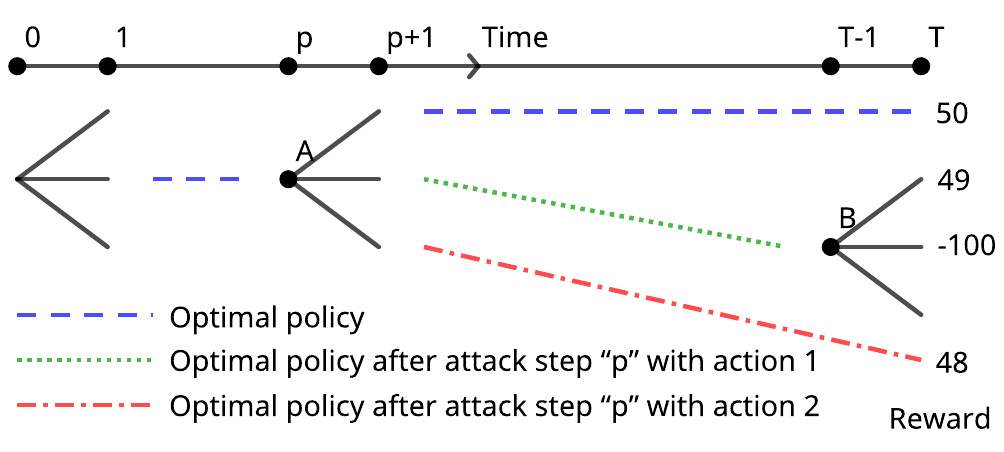}
		\caption{Example 2}
		\label{example2}
    \end{minipage}

\end{figure}

Consider a reward function like Figure \ref{example1}. All other terminal states have a reward in $[0,47]$.

For this game, if the sparse attack budget is not less than 2, the only optimal attack policy is attacking points B and C, resulting $-100$ reward.

But $Q_B^*(0)=50, Q_B^*(1)=49, Q_A^*(0)=50, Q_A^*(1)=48$. The $Q$ function at point B has less difference than point A. Therefore these two methods are both sub-optimal.

$\hfill\square$

Some other existing works used RL to learn the sparse attack timesteps. For example, \citet{behzadan2019adversarial} perturbed the action to $\arg\min_a Q(s_t,a)$ if the attacker decided to attack at timestep $t$; \citet{qiaoben2021strategically} used RL to learn the sparse attack timesteps with fixed adversary perturbation $h(s)$.

These methods could also be sub-optimal, because the learnt $Q$ function describes the optimal reward after having performed certain action, so perturbing action to $\arg\min_a Q$ only minimize the upper bound of the reward, not the lower bound. Here's an example:

\paragraph{Example 2}

The environment setting is the same as Example 1, except the action space is $\{0,1,2\}$. The game can be described by a ternary tree.

Consider a reward function like Figure \ref{example2}. All other terminal states have a reward in $[0,47]$.

For this game, if the sparse attack budget is not less than 2, the only optimal attack policy is attacking point A with action 1 and point B with action 1, resulting $-100$ reward.

But $Q_A^*(0)=50, Q_A^*(1)=49, Q_A^*(2)=48$, the optimal attack policy is not the one with the lowest $Q$ value. Therefore this method is sub-optimal.

$\hfill\square$

The detailed calculation of these two examples can be found in Appendix \ref{apdx1}

\subsection{Using (MA)RL algorithm to solve the optimal sparse attack policy}
\label{sec4.2}

The fundamental reason why rule-based sparse attacks are sub-optimal is that, the $Q$ function only describes optimal reward in certain case and it contains little information about the worst reward.
So in order to obtain the optimal sparse attack policy, one way is to use a (MA)RL algorithm to solve the regularized optimization problem.

Set $R'(s_t,\mathbf{a}_{\mathbf{k},t})=-R(s_t,\mathbf{a}_{\mathbf{k},t},\mathbf{a}_{-\mathbf{k},t}^*) {-} \lambda\sum_{i=1}^m 1_{[a_{k_i,t}\ne a_{k_i,t}^*]}$ and $P'(s_{t+1}|s_t,\mathbf{a}_{\mathbf{k},t})=P(s_{t+1}|s_t,\mathbf{a}_{\mathbf{k},t},\mathbf{a}_{-\mathbf{k},t}^*)$. Then the regularized optimization problem (\ref{eq:reg}) becomes a standard (M)MDP in the original environment, with agent $k_1,\cdots,k_m$, state transition function $P'$, and reward $R'$. Under some regularity assumptions, a optimal policy of the MDP exists \citep{puterman2014markov}, therefore we call this optimal policy as ``optimal sparse attack policy''.

This (M)MDP can be solved by any existing (MA)RL algorithm. 

For $m=1$, this is a partially observable single agent RL problem. Most common settings in partially observable RL assume that the true state is unknown. But in this case, the environment is still the original multi agent environment, in which the global state is known during the training period. We can take advantage of this feature and use the global state to help training. For example, we can apply the QMIX algorithm into this single agent case.

Algorithm \ref{alg:os} shows the learning procedure.

\begin{algorithm}[h]
	\caption{Optimal sparse attack}
	\label{alg:os}
	\begin{algorithmic}
		\REQUIRE Attack agents $k_1,\cdots,k_m$; Optimal Q function $Q_1^*,\cdots,Q_n^*$; Hyperparameter $\lambda$
		\ENSURE Optimal sparse attack Q function $Q_{k_1}^{adv},\cdots,Q_{k_m}^{adv}$
		\STATE Initialize replay buffer $\mathcal{D}=\emptyset$
		\FOR {each epoch}
		\FOR {sampling loop}
		\STATE Obtain current $s_t$ and $\mathbf{o}_t$. Compute optimal actions $a_{i,t}^*=\arg\max_a Q_i^*(o_{i,t},a)$ for $i=1,\cdots,n$.
		\STATE Roll out attacked agents' action $a_{{k_i},t}^{adv}$, using $\varepsilon$-greedy or other method, with $Q_{k_i}^{adv}(o_{{k_i},t},a)$ for $i=1,\cdots,m$
		\STATE Perform action $\mathbf{a}_{\mathbf{k},t}^{adv},\mathbf{a}_{-\mathbf{k},t}^*$ and get $r_t$, $s_{t+1}$ and $\mathbf{o}_{t+1}$.
		\STATE Compute attacked agents' reward $r'_t=-r_t-\lambda \sum_{i=1}^m 1_{[a_{{k_i},t}^{adv}\ne a_{{k_i},t}^*]}$
		\STATE Store transition $(s_t,\mathbf{o}_{\mathbf{k},t},\mathbf{a}_{\mathbf{k},t}^{adv},r'_t,s_{t+1},\mathbf{o}_{\mathbf{k},t+1})$ into $\mathcal{D}$
		\ENDFOR
		\FOR {training loop}
		\STATE Sample a minibatch $\mathcal{M}$ from replay buffer $\mathcal{D}$.
		\STATE Compute $Q_{k_i}^{adv}(o_{k_i},a_{k_i})$ for $i=1,\cdots,m$.
		\STATE Compute $Q(s,\mathbf{a}_\mathbf{k})$ using QMIX's mixer.
		\STATE Perform an update step.
		\ENDFOR
		\ENDFOR
		
	\end{algorithmic}
\end{algorithm}

\paragraph{Comparison with existing works.}
Except for sub-optimal rule based sparse attack discussed earlier, \citet{sun2020stealthy} used RL to learn the optimal sparse attack either, and \citet{lin2020robustness,guo2022towards,pham2022evaluating} used RL to learn the optimal attack in cMARL system. The main difference between our work and theirs is:
\begin{itemize}
	\item \citet{sun2020stealthy} only considered attacking single agent RL policy. Our work focuses on attacking multi agent RL policy, and testing on many different environments. Their work used hard constrain applied to the environment to control the sparsity, i.e., the attacker will fail to attack if maximum of attack step reached. Our work uses soft regularization to control the sparsity.
	\item Both \citet{lin2020robustness,guo2022towards,pham2022evaluating} considered the dense attack, and the first two works only attempted to attack one agent in the team. Our work focuses on sparse attack, and attempts to attack both single agent and multiple agents.
\end{itemize}

\section{Experiments}
\label{sec5}

\subsection{Experiment settings}
\label{sec5.1}

We evaluate our optimal sparse attack method on SMAC environment \cite{samvelyan2019starcraft} with different maps. For maps with only a few agents, we only attempt to attack a single agent; For maps with many agents, we attempt to attack both single agent and multiple agents.

\paragraph{Environment and experiment settings}

SMAC \cite{samvelyan2019starcraft} is an MIT-licensed multi agent partially observable environment based on the StarCraft II game. Each agent can only observe a part of the global state within a circular area around it. Please refer to the SMAC paper for more details.

We use QMIX \cite{rashid2018qmix} to train the base policy to be attacked. We mainly focus on attacking QMIX policies, but we also try attacking VDN \cite{sunehag2017value} and QTRAN \cite{son2019qtran} policies in some maps. The base policy is trained with the same settings and same parameters as in the QMIX paper. 
We also use QMIX to train the optimal sparse attack policy. The optimal sparse attack policy is also trained with the same settings and same parameters as in the QMIX paper, except the number of agents is changed.

\paragraph{Map selection}

SMAC environment contains many different maps, but only those maps that QMIX can achieve good performance are suitable for evaluating the performance of the sparse attack policy.
We evaluate our method on different kinds of maps:

\begin{itemize}
	\item Maps with a few agents: 3m and 3s\_vs\_4z. In these two maps, we only attack single agent. Because all the agents are homogeneous, we just choose agent 0 as a representative agent. 
	\item Maps with a medium number of agents: 2s3z, 8m, 1c3s5z, so\_many\_baneling. In these maps, we attempt to attack both single agent and multiple agents. For homogeneous map 8m and so\_many\_baneling, we just randomly choose the representative agent; for heterogeneous map, we choose the representative agent based on agent type.
	\item Maps with a large number of agents: 25m and bane\_vs\_bane. In these maps, we attempt to attack both single agent and multiple agents. Since the number of agents is large, we just randomly choose the representative agent.
\end{itemize}

Some detailed explanation can be found in Appendix \ref{apdx2}.

\paragraph{Evaluation criteria and baselines}

We use the winning rate of 1000 episodes and the average number of attack steps as our evaluation criteria. The attack policy with fewer attack steps and a lower winning rate has better performance.

To evaluate the stability of the experiment, we run each (MA)RL training 5 times with different randomly selected random seeds, discard the policy with the highest and lowest winning rate, and report the result of the rest 3 policies.

We evaluate our result with these baselines:

\begin{itemize}
	\item \textbf{Ra}ndom sparse attack. At each timesteps we randomly attack it with a certain probability, with either \textbf{r}andom action, or use the action that have the \textbf{l}owest Q value. We use the abbreviation Ra-R or Ra-L to represent them.
	\item \textbf{Ru}le \textbf{b}ased sparse attack. Following \citet{lin2017tactics}, we compute $\delta=\max_a \text{softmax}(Q)-\min_a \text{softmax}(Q)$ and attack those timesteps with high $\delta_t$ and with the action that have the lowest Q value. We use the abbreviation Ru-B to represent it.
	\item \textbf{RL} with \textbf{f}ixed attack policy. Following \citet{behzadan2019adversarial}, we use RL to learn the attack timesteps, and then attack those timesteps with the action that have the lowest Q value. We use the abbreviation RL-F to represent it. In \citet{behzadan2019adversarial} they fixed the attack cost $c_{adv}$ to $1$ per attack step, but in order to make these result comparable, we may adjust $c_{adv}$ to control the attack sparsity.
	\item \textbf{Ru}le based \textbf{d}ense attack. Attack every timesteps with the action that have the lowest Q value. This baseline is only used in a few case, because it is a much stronger attack. We use the abbreviation Ru-D to represent it.
\end{itemize}

We use the abbreviation OPT to represent our method: \textbf{opt}imal sparse attack.
For hyperparameter $\lambda$, we train different attack policies with different $\lambda$, and choose the one that can significantly lower the winning rate. Results below show that larger $\lambda$ leads to less attack.

\subsection{Maps with a few agents}
\label{sec5.2}

Table \ref{tab:fa-rst} shows the result of the 3m map and the 3s\_vs\_4z map. For these two maps, the winning rate without adversarial attack is 97.2\% and 90.8\% respectively. We can see that, for 3m map, OPT can lower the winning rate to 0\% with only around 18\% of timesteps being attacked, and 1.8\% with around 14\%. For 3s\_vs\_4z map, the winning rate cannot be as low as 3m map, but OPT can still significantly lower the winning rate with around 10\%-15\% of timesteps being attacked.

These results show that OPT outperforms Ra-R/Ra-L and Ru-B. In 3m map, the performance of OPT and RL-F is similar, but in 3s\_vs\_4z OPT significantly outperforms RL-F. Note that, for Ru-B and RL-F, we do not need to accurately select a threshold or $c_{adv}$ to let them attack exactly the same number of timesteps as OPT. If OPT attacks fewer timesteps than Ru-B or RL-F, and achieves a lower winning rate, then OPT just outperforms them.

\begin{table}
	\caption{Maps with a few agents}
	\label{tab:fa-rst}
	\begin{center}
	\begin{tabular}{c|ccc|ccc}
		\hline
		Map & \multicolumn{3}{c|}{3m Agent 0} & \multicolumn{3}{c}{3s\_vs\_4z Agent 0} \\\hline
		\begin{tabular}[c]{@{}c@{}}Attack\\ type\end{tabular} & Parameter & \begin{tabular}[c]{@{}c@{}}Winning\\ rate (\%)\end{tabular} & \begin{tabular}[c]{@{}c@{}}Attacked steps / \\ Total steps\end{tabular} & Parameter & \begin{tabular}[c]{@{}c@{}}Winning\\ rate (\%)\end{tabular} & \begin{tabular}[c]{@{}c@{}}Attacked steps / \\ Total steps\end{tabular} \\ \hline
		\multirow{6}{*}{OPT} & \multirow{3}{*}{$\lambda=1$} & 0.0 & 4.278/23.249 & \multirow{3}{*}{$\lambda=0.1$} & 56.5 & 10.123/167.604 \\
		& & 0.0 & 4.336/23.067 & & 58.3 & 15.993/171.329 \\
		& & 0.0 & 4.329/23.356 & & 59.9 & 9.404/165.055 \\ \cline{2-7} 
		& \multirow{3}{*}{$\lambda=5$} & 1.8 & 3.464/24.906  & \multirow{3}{*}{$\lambda=0.01$} & 46.0 & 22.085/163.402  \\
		& & 1.4 & 3.640/24.975  & & 49.0 & 24.480/175.448  \\
		& & 1.4 & 3.460/25.436  & & 46.7 & 20.370/167.866  \\ \hline
		\multirow{2}{*}{Ra-R} & \multirow{2}{*}{-} & 85.3 & 20\% & \multirow{2}{*}{-} & 90.5 & 15\% \\
		& & 88.4 & 15\% & & 91.6 & 10\% \\ \hline
		\multirow{2}{*}{Ra-L} & \multirow{2}{*}{-} & 48.6 & 20\% & \multirow{2}{*}{-} & 87.2 & 15\% \\
		& & 60.3 & 15\% & & 90.5 & 10\% \\ \hline
		\multirow{2}{*}{Ru-B} & \begin{tabular}[c]{@{}c@{}}Threshold\\ 0.25\end{tabular} & 26.7 & \begin{tabular}[c]{@{}c@{}}15.657/39.389\\ \end{tabular} & \begin{tabular}[c]{@{}c@{}}Threshold\\ 0.04\end{tabular} & 66.2 & \begin{tabular}[c]{@{}c@{}}30.794/159.433\\ \end{tabular} \\
		& \begin{tabular}[c]{@{}c@{}}Threshold\\ 0.258\end{tabular} & 41.1 & \begin{tabular}[c]{@{}c@{}}12.475/35.468\\ \end{tabular} & \begin{tabular}[c]{@{}c@{}}Threshold\\ 0.05\end{tabular} & 90.4 & \begin{tabular}[c]{@{}c@{}}19.811/136.088\\ \end{tabular} \\ \hline
		RL-F & $c_{adv}=1$ & 0.0 & 3.997/40.113 & $c_{adv}=0.1$ & 78.6 & 15.958/150.592 \\ \hline
	\end{tabular}
\end{center}
\end{table}

	\subsection{Maps with a medium number of agents}
\label{sec5.3}

For 8m, 2s3z, 1c3s5z, so\_many\_baneling map, we attempt to attack both single agent and multiple agents. The winning rate of these 4 maps without adversarial attack is 87.9\%, 97.5\%, 99.5\%, 95.0\% respectively.

\paragraph{Attacking single agent}

We pick one agent for each map to report here, and more experimental results can be found in Appendix \ref{apdx3}.

Table \ref{tab:med-sig-rst} shows the result of these 4 maps. OPT can lower the winning rate in all these 4 maps. Especially in 1c3s5z, OPT lowers the winning rate to less than 1\% in only less than 10\% of timesteps being attacked. 

In both maps, OPT outperforms Ra-R, Ra-L and Ru-B. In 8m map, the performance of OPT and RL-F is similar, and in the other 3 maps, OPT significantly outperforms RL-F.

\begin{table}[]
	\caption{Attacking single agent}
	\label{tab:med-sig-rst}
	\begin{tabular}{c|ccc|ccc}
	\hline
	Map                                                   & \multicolumn{3}{c|}{8m Agent 4}                                                                                                                                                                 & \multicolumn{3}{c}{so\_many\_baneling Agent 6}                                                                                                                                                  \\ \hline
	\begin{tabular}[c]{@{}c@{}}Attack\\ type\end{tabular} & Parameter                                                & \begin{tabular}[c]{@{}c@{}}Winning\\ rage (\%)\end{tabular} & \begin{tabular}[c]{@{}c@{}}Attacked steps /\\ Total steps\end{tabular} & Parameter                                                & \begin{tabular}[c]{@{}c@{}}Winning\\ rage (\%)\end{tabular} & \begin{tabular}[c]{@{}c@{}}Attacked steps /\\ Total steps\end{tabular} \\ \hline
	\multirow{3}{*}{OPT}                                  & \multirow{3}{*}{$\lambda=1$}                             & 0.2                                                         & 7.109/26.876                                                           & \multirow{3}{*}{$\lambda=0.5$}                           & 8.4                                                         & 4.688/24.229                                                           \\
														  &                                                          & 0.2                                                         & 7.153/27.393                                                           &                                                          & 6.6                                                         & 4.720/24.259                                                           \\
														  &                                                          & 0.1                                                         & 5.008/26.373                                                           &                                                          & 7.9                                                         & 4.711/24.054                                                           \\ \hline
	Ra-R                                                  & -                                                        & 71.4                                                        & 25\%                                                                   & -                                                        & 89.8                                                        & 25\%                                                                   \\ \hline
	Ra-L                                                  & -                                                        & 33.7                                                        & 25\%                                                                   & -                                                        & 87.1                                                        & 25\%                                                                   \\ \hline
	Ru-B                                                  & \begin{tabular}[c]{@{}c@{}}Threshold\\ 0.13\end{tabular} & 24.1                                                        & 8.291/33.548                                                           & \begin{tabular}[c]{@{}c@{}}Threshold\\ 0.03\end{tabular} & 78.0                                                        & 11.440/24.830                                                          \\ \hline
	RL-F                                                  & $c_{adv}=0.25$                                           & 0.6                                                         & 6.431/99.718                                                           & $c_{adv}=0.1$                                            & 53.6                                                        & 7.245/25.194                                                           \\ \hline
	Map                                                   & \multicolumn{3}{c|}{1c3s5z Agent 0}                                                                                                                                                             & \multicolumn{3}{c}{2s3z Agent 1}                                                                                                                                                                \\ \hline
	\begin{tabular}[c]{@{}c@{}}Attack\\ type\end{tabular} & Parameter                                                & \begin{tabular}[c]{@{}c@{}}Winning\\ rage (\%)\end{tabular} & \begin{tabular}[c]{@{}c@{}}Attacked steps /\\ Total steps\end{tabular} & Parameter                                                & \begin{tabular}[c]{@{}c@{}}Winning\\ rage (\%)\end{tabular} & \begin{tabular}[c]{@{}c@{}}Attacked steps /\\ Total steps\end{tabular} \\ \hline
	\multirow{6}{*}{OPT}                                  & \multirow{3}{*}{$\lambda=1$}                             & 0.0                                                         & 4.097/44.122                                                           & \multirow{3}{*}{$\lambda=0.1$}                           & 0.4                                                         & 11.413/119.696                                                         \\
														  &                                                          & 0.1                                                         & 4.134/44.389                                                           &                                                          & 0.5                                                         & 10.942/119.577                                                         \\
														  &                                                          & 0.0                                                         & 4.019/45.706                                                           &                                                          & 0.5                                                         & 11.528/119.454                                                         \\ \cline{2-7} 
														  & \multirow{3}{*}{$\lambda=5$}                             & 0.3                                                         & 3.617/46.609                                                           & \multirow{3}{*}{$\lambda=1$}                             & 7.2                                                         & 11.116/113.693                                                         \\
														  &                                                          & 0.3                                                         & 3.458/45.725                                                           &                                                          & 2.2                                                         & 10.062/118.165                                                         \\
														  &                                                          & 0.3                                                         & 3.452/45.555                                                           &                                                          & 2.5                                                         & 10.193/117.992                                                         \\ \hline
	Ra-R                                                  & -                                                        & 96.8                                                        & 10\%                                                                   & -                                                        & 97.1                                                        & 10\%                                                                   \\ \hline
	Ra-L                                                  & -                                                        & 94.7                                                        & 10\%                                                                   & -                                                        & 91.1                                                        & 10\%                                                                   \\ \hline
	Ru-B                                                  & \begin{tabular}[c]{@{}c@{}}Threshold\\ 0.1\end{tabular}  & 45.7                                                        & 5.83/64.444                                                            & \begin{tabular}[c]{@{}c@{}}Threshold\\ 0.1\end{tabular}  & 50.8                                                        & 12.709/56.991                                                          \\ \hline
	RL-F                                                  & $c_{adv}=0.3$                                            & 2.5                                                         & 4.602/48.968                                                           & $c_{adv}=0.1$                                            & 22.1                                                        & 32.688/88.48                                                           \\ \hline
	\end{tabular}
	\end{table}

\paragraph{Attacking multiple agents}

Table \ref{tab:med-mul-rst} shows the result of multiple agent attack, and additional results can be found in Appendix \ref{apdx3}. In these maps, OPT can significantly lower the winning rate, and outperforms all baselines including RL-F. In 1c3s5z map, OPT outperforms Ru-D. In maps with heterogeneous agents, different types of agents contribute differently to the team, small agents play less of a role in the team. Compared to previous result, in 1c3s5z map, ``c'' is large agent and ``s'' / ``z'' are small agents, so attacking ``s'' / ``z'' requires more attack steps than attacking ``c''.

\begin{table}[]
	\caption{Attacking multiple agents}
	\label{tab:med-mul-rst}
	\begin{tabular}{c|ccc|ccc}
	\hline
	Map                                                   & \multicolumn{3}{c|}{so\_many\_baneling Agent 0 and 6}                                                                                                                                                & \multicolumn{3}{c}{1c3s5z Agent 1 and 7}                                                                                                                                                              \\ \hline
	\begin{tabular}[c]{@{}c@{}}Attack\\ type\end{tabular} & Parameter                                                     & \begin{tabular}[c]{@{}c@{}}Winning\\ rage (\%)\end{tabular} & \begin{tabular}[c]{@{}c@{}}Attacked steps /\\ Total steps\end{tabular} & Parameter                                                      & \begin{tabular}[c]{@{}c@{}}Winning\\ rage (\%)\end{tabular} & \begin{tabular}[c]{@{}c@{}}Attacked steps /\\ Total steps\end{tabular} \\ \hline
	\multirow{6}{*}{OPT}                                  & \multirow{3}{*}{$\lambda=0.5$}                                & 3.7                                                         & \begin{tabular}[c]{@{}c@{}}5.643,1.204\\ /21.219\end{tabular}          & \multirow{3}{*}{$\lambda=0.1$}                                 & 48.3                                                        & \begin{tabular}[c]{@{}c@{}}16.265,13.545\\ /106.380\end{tabular}       \\
														  &                                                               & 2.3                                                         & \begin{tabular}[c]{@{}c@{}}5.755,5.150\\ /21.199\end{tabular}          &                                                                & 49.6                                                        & \begin{tabular}[c]{@{}c@{}}15.507,15.621\\ /110.040\end{tabular}       \\
														  &                                                               & 3.3                                                         & \begin{tabular}[c]{@{}c@{}}5.988,1.344\\ /21.276\end{tabular}          &                                                                & 48.2                                                        & \begin{tabular}[c]{@{}c@{}}15.729,14.188\\ /112.490\end{tabular}       \\ \cline{2-7} 
														  & \multirow{3}{*}{$\lambda=1$}                                  & 6.4                                                         & \begin{tabular}[c]{@{}c@{}}5.405,1.037\\ /21.044\end{tabular}          & \multirow{3}{*}{$\lambda=0.01$}                                & 29.1                                                        & \begin{tabular}[c]{@{}c@{}}57.834,58.535\\ /79.101\end{tabular}        \\
														  &                                                               & 8.1                                                         & \begin{tabular}[c]{@{}c@{}}5.423,1.015\\ /21.299\end{tabular}          &                                                                & 29.7                                                        & \begin{tabular}[c]{@{}c@{}}53.077,56.998\\ /78.544\end{tabular}        \\
														  &                                                               & 8.9                                                         & \begin{tabular}[c]{@{}c@{}}5.600,1.090\\ /21.200\end{tabular}          &                                                                & 24.9                                                        & \begin{tabular}[c]{@{}c@{}}54.735,48.304\\ /78.564\end{tabular}        \\ \hline
	Ra-R                                                  & -                                                             & 90.0                                                        & 30\%, 10\%                                                             & -                                                              & 97.9                                                        & 15\%, 15\%                                                             \\ \hline
	Ra-L                                                  & -                                                             & 84.8                                                        & 30\%, 10\%                                                             & -                                                              & 98.3                                                        & 15\%, 15\%                                                             \\ \hline
	Ru-D                                                  & -                                                             & -                                                           & -                                                                      & -                                                              & 37.7                                                        & 100\%, 100\%                                                           \\ \hline
	Ru-B                                                  & \begin{tabular}[c]{@{}c@{}}Threshold\\ 0.05, 0.1\end{tabular} & 74.4                                                        & \begin{tabular}[c]{@{}c@{}}11.714,2.460\\ /29.005\end{tabular}         & \begin{tabular}[c]{@{}c@{}}Threshold\\ 0.04, 0.05\end{tabular} & 68.8                                                        & \begin{tabular}[c]{@{}c@{}}23.486,27.912\\ /73.038\end{tabular}        \\ \hline
	RL-F                                                  & $c_{adv}=0.1$                                                 & 17.5                                                        & \begin{tabular}[c]{@{}c@{}}4.783,6.286\\ /25.019\end{tabular}          & $c_{adv}=0.01$                                                 & 37.0                                                        & \begin{tabular}[c]{@{}c@{}}77.475,85.843\\ /118.839\end{tabular}       \\ \hline
	\end{tabular}
	\end{table}

	\subsection{Maps with a large number of agents}
\label{sec5.4}

For 25m and bane\_vs\_bane map, we also attempt to attack both single agent and multiple agents. 

Table \ref{tab:mass-rst} shows the result of the 25m map. The winning rate of the 25m map without adversarial attack is 98.4\%. More results can be found in Appendix \ref{apdx3}. 

When attacking only one agent in a team with 25 agents, OPT can lower the winning rate a little bit with around 80\% of timesteps being attacked. This still outperforms Ru-D and RL-F. Ru-B has no baseline in this case, because we fail to find a threshold that can attack more than 80\% but not near 100\% of timesteps.
When attacking 5 agents in the team, OPT can significantly lower the winning rate with each agent around 7\% of timesteps being attacked. All these results outperform all baselines including RL-F.

\begin{table}
	\caption{25m map}
	\label{tab:mass-rst}
	\begin{center}

				\begin{tabular}{c|ccc}
					\hline
					Map & \multicolumn{3}{c}{25m Agent 10} \\ \hline
					\begin{tabular}[c]{@{}c@{}}Attack\\ type\end{tabular} & Parameter & \begin{tabular}[c]{@{}c@{}}Winning\\ rate (\%)\end{tabular} & \begin{tabular}[c]{@{}c@{}}Attacked steps / \\ Total steps\end{tabular} \\ \hline
					\multirow{6}{*}{OPT} & \multirow{3}{*}{$\lambda=0.01$} & 69.3 & 30.302/46.582  \\
					& & 68.1 & 37.845/48.739  \\
					& & 70.8 & 44.589/56.135  \\ \cline{2-4} 
					& \multirow{3}{*}{$\lambda=0.001$} & 67.0 & 51.887/63.311  \\
					& & 73.8 & 45.496/56.599  \\
					& & 65.7 & 49.720/63.385  \\ \hline
					Ra-R & - & 83.7 & 70\% \\ \hline
					Ra-L & - & 73.8 & 70\% \\\hline
					Ru-D & - & 71.8 & 100\% \\ \hline
					RL-F & $c_{adv}=0$ & 70.7 & 42.459/62.132 \\ \hline
					Map & \multicolumn{3}{c}{25m Agent 0,5,10,15,20} \\ \hline
					\begin{tabular}[c]{@{}c@{}}Attack\\ type\end{tabular} & Parameter & \begin{tabular}[c]{@{}c@{}}Winning\\ rate (\%)\end{tabular} & \begin{tabular}[c]{@{}c@{}}Attacked steps / \\ Total steps\end{tabular} \\ \hline
					\multirow{3}{*}{OPT} & \multirow{3}{*}{$\lambda=1$} & 16.2 & 3.700,3.840,3.567,4.051,3.467/58.801 \\
					& & 16.2 & 3.575,3.928,3.651,3.999,3.484/59.101  \\
					& & 19.3 & 3.938,3.715,3.608,3.881,3.465/56.315  \\ \hline
					Ra-R & - & 96.5 & 7\%, 7\%, 7\%, 7\%, 7\% \\ \hline
					Ra-L & - & 89.4 & 7\%, 7\%, 7\%, 7\%, 7\% \\ \hline
					Ru-B & \begin{tabular}[c]{@{}c@{}}Threshold \\ 0.16,0.13,0.11,0.13,0.15\end{tabular} & 42.8 & 6.729,4.721,6.189,6.855,5.341/50.164\\ \hline
					RL-F & $c_{adv}=0.2$ & 35.9 & 7.335,36.529,6.959,2.511,2.530/101.708 \\ \hline
				\end{tabular}

	\end{center}

\end{table}

\subsection{Attacking VDN and QTRAN policies}

We also attack VDN and QTRAN policies in two maps: 1c3s5z and 25m, i.e., one homogeneous map and one heterogeneous map. Results can be found in Appendix \ref{apdx3}. The result is similar to QMIX policies. Both VDN and QTRAN policies can perform awful under sparse attack, and OPT outperforms all baseline methods.

	\section{Conclusion and Future works}
	\label{sec6}

In this work we have been concerned with sparse attack in multi agent reinforcement learning. We have used (MA)RL algorithms to learn the optimal sparse attack policy and evaluated them in SMAC environments. In most of our experiments, our method outperformed all baselines, which indicated that rule based sparse attack method is sub-optimal. Our experiments have shown that the policy trained by the QMIX / VDN / QTRAN algorithm can obtain poor performance when only one or a few agents in the team are attacked at a few timesteps. 

Our results implied that existing cMARL algorithms, at least QMIX / VDN / QTRAN, may not be robust enough to deploy to real applications. So one straightforward future work is to find some ways to defend this kind of attack. To the best of our knowledge, this work is the first work about sparse attack in MARL. As a first-step work, we fixed the target agent and then learn the optimal sparse attack policy. A more general attack problem is to take $k_1,\cdots,k_m$ into the $\min$ operator, i.e., the attacker must decide both the agent to be attacked and the attack policy. This problem is a mixed integer programming problem, and is also a possible future work.


\bibliography{neurips_2022}

\bibliographystyle{plainnat}



\appendix

\section{Appendix}
\label{appendix}

\subsection{Detailed derivation}
\label{apdx1}

\paragraph{Example 1}

Since the game can be described by a binary tree, we can just use action sequence to denote each state, i.e., action sequence $a_0,a_1,\cdots,a_{t-1}$ will lead to state $s_{a_0,a_1,\cdots,a_{t-1}}$ at timestep $t$. The reward function at terminal step $T$ can be denoted by $r_{a_0,a_1,\cdots,a_{T-1}}$.

Consider this reward function:
\begin{itemize}
	\item Choose an action sequence $a_0^*,\cdots,a_{T-1}^*$ as optimal policy, and set $r_{a_0^*,\cdots,a_{T-1}^*}=50$
	\item Choose some $p<t<T-1$. Choose another action sequence $a_{t+1}',\cdots,a_{T-2}'$, and set $r_{a_0^*,\cdots,a_{t-1}^*,1-a_{t}^*,a_{t+1}',\cdots,a_{T-2}',1}=49$, $r_{a_0^*,\cdots,a_{t-1}^*,1-a_{t}^*,a_{t+1}',\cdots,a_{T-2}',0}=-100$
	\item Choose another action sequence $a_{p+1}'',\cdots,a_{T-1}''$, and set $r_{a_1^*,\cdots,a_{p-1}^*,1-a_{p}^*,a_{p+1}'',\cdots,a_{T-1}''}=48$
	\item For any other terminal state, set its reward to any value between $[0,47]$
\end{itemize}

For this game, if the sparse attack budget is not less than 2, the only optimal attack policy is to attack step $t$ and step $T-1$, with reward -100:
\begin{itemize}
	\item $r_{a_0^*,\cdots,a_{t-1}^*,1-a_{t}^*,a_{t+1}',\cdots,a_{T-2}',1}=49$, and any other terminal state's reward is at most $48$. If we first attack step $t$ and change its action to $1-a_t^*$, then the rest optimal policy is $a_{t+1}',\cdots,a_{T-2}',1$.
	\item Then we attack step $T-1$ and change its action from $1$ to $0$, agent will receive $-100$ reward, which is the only worst reward of the whole game (Any other terminal state's reward is at least $0$).
\end{itemize}

Now let's calculate some $Q^*$ function: 
\begin{itemize}
	\item At timestep $t$, $Q_t^*(s_{a_0^*,\cdots,a_{t-1}^*},a_t^*)=50$, $Q_t^*(s_{a_0^*,\cdots,a_{t-1}^*},1-a_t^*)=49$ (The rest optimal policy is $a_{t+1}',\cdots,a_{T-2}',1$, which will receive $49$ reward)
	\item At timestep $p$, $Q_p^*(s_{a_0^*,\cdots,a_{p-1}^*},a_p^*)=50$, $Q_p^*(s_{a_0^*,\cdots,a_{p-1}^*},1-a_p^*)=48$ (The rest optimal policy is $a_{p+1}'',\cdots,a_{T-1}''$, which will receive $48$ reward. Any other policy can only get at most $47$ reward.)
\end{itemize}

Now, no matter what rule we choose, $\delta_p$ is larger than $\delta_t$, which means that the rule based sparse attack using $\delta$ is sub-optimal. Since the optimal attack policy requires attack at $t$ while do not attack at $p$, which is not possible.

\paragraph{Example 2}

Similar to Example 1, we use action sequence to denote each state and reward function.

Consider this reward function:
\begin{itemize}
	\item Choose an action sequence $a_0^*,\cdots,a_{T-1}^*$ as optimal policy, and set $r_{a_0^*,\cdots,a_{T-1}^*}=50$
	\item Choose some $p<T-1$. Choose another action sequence $a_{p+1}',\cdots,a_{T-2}'$, and set $r_{a_0^*,\cdots,a_{p-1}^*,(a_p^*+1)\text{ mod } 3,a_{p+1}',\cdots,a_{T-2}',1}=49$, $r_{a_0^*,\cdots,a_{p-1}^*,(a_p^*+1)\text{ mod } 3,a_{p+1}',\cdots,a_{T-2}',0}=-100$
	\item Choose another action sequence $a_{p+1}'',\cdots,a_{T-1}''$, and set $r_{a_0^*,\cdots,a_{p-1}^*,(a_p^*+2)\text{ mod } 3,a_{p+1}'',\cdots,a_{T-1}''}=48$
	\item For any other terminal state, set its reward to any value between $[0,47]$
\end{itemize}

For this game, if the sparse attack budget is not less than 2, the only optimal attack policy is to attack step $p$ and step $T-1$, with reward -100:
\begin{itemize}
	\item $r_{a_0^*,\cdots,a_{p-1}^*,(a_p^*+1)\text{ mod } 3,a_{p+1}',\cdots,a_{T-2}',1}=49$, and any other terminal state's reward is at most $48$. If we first attack step $p$ and change its action to $(a_p^*+1)\text{ mod } 3$, then the rest optimal policy is $a_{p+1}',\cdots,a_{T-2}',1$
	\item Then we attack step $T-1$ and change its action to $0$, agent will receive $-100$ reward, which is the only worst reward of the whole game (Any other terminal state's reward is at least $0$)
\end{itemize}

Now let's calculate some $Q^*$ function: 
\begin{itemize}
	\item $Q_p^*(s_{a_0^*,\cdots,a_{p-1}^*},a_p^*)=50$
	\item $Q_p^*(s_{a_0^*,\cdots,a_{p-1}^*},(a_p^*+1)\text{ mod } 3)=49$ (The rest optimal policy is $a_{p+1}',\cdots,a_{T-2}',1$, which will receive $49$ reward)
	\item $Q_p^*(s_{a_0^*,\cdots,a_{p-1}^*},(a_p^*+2)\text{ mod } 3)=48$ (The rest optimal policy is $a_{p+1}'',\cdots,a_{T-1}''$, which will receive $48$ reward. Any other policy can only get at most $47$ reward.)
\end{itemize}

Now, if the agent decide to attack at step $p$ with no attack before, the optimal attack action $(a_p^*+1)\text{ mod } 3$ is not the one that has the lowest $Q$ value.

\subsection{Map selection}
\label{apdx2}

The following factors are considered when choosing maps:
\begin{itemize}
	\item The performance of the QMIX algorithm. If QMIX itself performs poorly in a map, then there is little value in performing an adversarial attack on it.
	\item Both homogeneous maps and heterogeneous maps should be chosen.
	\item The difficulty of the map. Since the more difficult the map itself, the easier for adversarial attack. Therefore we prefer simple maps than difficult maps.
\end{itemize}

Also, since we focus on attacking multi agent systems, we do not choose the environment with only two agents.

The first factor excludes 5m\_vs\_6m, MMM2, 3s5z\_vs\_3s6z, corridor, 6h\_vs\_8z. See QMIX paper for specific.

For homogeneous maps, we choose 3m, 8m, 25m, 3s\_vs\_4z, and so\_many\_baneling. We do not choose 8m\_vs\_9m, 10m\_vs\_11m, or 27m\_vs\_30m, since they are more difficult than the chosen ``m'' maps because the team has fewer agents than the enemy. The 3s\_vs\_{3,4,5}z maps are similar, so we just choose one of them.

All other maps are heterogeneous. 2s3z, 3s5z, and 1c3s5z have some similarity (both contains ``s'' and ``z'' agent), and we choose 2s3z and 1c3s5z. For the rest two maps, we choose bane\_vs\_bane since it is the only heterogeneous map with a large number of agents.

\subsection{Additional experiment results}
\label{apdx3}

Table \ref{tab:2s3z-rst} shows additional experiment results of the 2s3z map. OPT can lower the winning rate to less than 10\% in both cases. All results outperform all baselines.

\begin{table}
	\caption{2s3z map}
	\label{tab:2s3z-rst}
	\begin{center}

				\begin{tabular}{c|ccc|ccc}
					\hline
					Map & \multicolumn{3}{c|}{2s3z Agent 2} & \multicolumn{3}{c}{2s3z Agent 1 and 2} \\ \hline
					\begin{tabular}[c]{@{}c@{}}Attack\\ type\end{tabular} & Parameter & \begin{tabular}[c]{@{}c@{}}Winning\\ rate (\%)\end{tabular} & \begin{tabular}[c]{@{}c@{}}Attacked steps / \\ Total steps\end{tabular} & Parameter & \begin{tabular}[c]{@{}c@{}}Winning\\ rate (\%)\end{tabular} & \begin{tabular}[c]{@{}c@{}}Attacked steps / \\ Total steps\end{tabular} \\ \hline
					\multirow{6}{*}{OPT} & \multirow{3}{*}{$\lambda=0.1$} & 0.0 & 21.715/44.057 & \multirow{3}{*}{$\lambda=0.1$} & 0..0 & 17.782,16.216/43.321  \\
					& & 0.0 & 24.301/46.361  & & 0.0 & 17.294,19.044/44.120  \\
					& & 0.0 & 22.063/43.904 & & 0.0 & 17.501,16.134/43.904  \\ \cline{2-7} 
					& \multirow{3}{*}{$\lambda=0.5$} & 11.1 & 12.547/48.199  & \multirow{3}{*}{$\lambda=0.5$} & 0.4 & 9.142,8.593/81.301 \\
					& & 1.5 & 13.875/46.908  & & 0.9 & 6.161,12.821/48.101  \\
					& & 5.0 & 13.363/49.769  & & 0.1 & 7.277,12.651/44.000  \\ \hline
					Ra-R & - & 91.3 & 15\% & - & 76.0 & 10\%, 30\% \\
					& - & 41.7 & 50\% & & & \\ \hline
					Ra-L & - & 84.5 & 15\% & - & 39.5 & 10\%, 30\% \\
					& - & 5.4 & 50\% & & & \\ \hline
					Ru-B & \begin{tabular}[c]{@{}c@{}}Threshold\\ 0.07\end{tabular} & 7.2 & 15.052/52.341  & \begin{tabular}[c]{@{}c@{}}Threshold\\ 0.087,0.115\end{tabular} & 3.9 & 10.030,12.258/47.077  \\ \hline
					RL-F & $c_{adv}=0.2$ & 17.6 & 16.584/76.338 & $c_{adv}=0.1$ & 12.3 & 19.444,10.877/96.483 \\ \hline
				\end{tabular}

	\end{center}
\end{table}

Table \ref{tab:1c3s5z-rst} shows additional experiment results of the 1c3s5z map. In this map, ``c'' is a large agent and ``s''/``z'' are small agents, therefore ``c'' is more important to the team. Compared to previous results, these results show that attacking ``s'' (agent 1) or ``z'' (agent 7) is more difficult than attacking ``c'' (agent 0). For agent 7, OPT outperforms all baselines including Ru-D and RL-F. For agent 1, the result is just so-so, but with $\lambda=0.1$ OPT still outperforms Ru-B and RL-F. 

\begin{table}
	\caption{1c3s5z map}
	\label{tab:1c3s5z-rst}

	\begin{center}
				\begin{tabular}{c|ccc|ccc}
					\hline
					Map & \multicolumn{3}{c|}{1c3s5z Agent 7} & \multicolumn{3}{c}{1c3s5z Agent 1} \\ \hline
					\begin{tabular}[c]{@{}c@{}}Attack\\ type\end{tabular} & Parameter & \begin{tabular}[c]{@{}c@{}}Winning\\ rate (\%)\end{tabular} & \begin{tabular}[c]{@{}c@{}}Attacked steps / \\ Total steps\end{tabular} & Parameter & \begin{tabular}[c]{@{}c@{}}Winning\\ rate (\%)\end{tabular} & \begin{tabular}[c]{@{}c@{}}Attacked steps / \\ Total steps\end{tabular} \\ \hline
					\multirow{6}{*}{OPT} & \multirow{3}{*}{$\lambda=0.1$} & 92.6 & 10.407/48.971 & \multirow{3}{*}{$\lambda=0.1$} & 83.6 & 13.408/69.093  \\
					& & 92.7 & 10.907/48.597 & & 83.8 & 14.504/68.233  \\
					& & 92.5 & 9.013/49.212  & & 84.2 & 14.731/65.389 \\ \cline{2-7} 
					& \multirow{3}{*}{$\lambda=0.01$} & 78.6 & 42.252/60.875  & \multirow{3}{*}{$\lambda=0.01$} & 84.2 & 48.083/60.750  \\
					& & 79.7 & 42.170/55.780  & & 86.4 & 45.741/59.606  \\
					& & 77.1 & 45.324/58.568  & & 79.8 & 51.944/67.537 \\ \hline
					Ra-R & - & 98.0 & 20\% & - & 98.1 & 20\% \\
					& & & & & 96.9 & 80\% \\ \hline
					Ra-L & - & 98.5 & 20\% & - & 98.6 & 20\% \\
					& & & & & 87.8 & 80\% \\ \hline
					Ru-D & - & 89.5 & 100\% & - & 78.4 & 100\% \\ \hline
					Ru-B & \begin{tabular}[c]{@{}c@{}}Threshold\\ 0.05\end{tabular} & 96.9 & 16.756/48.097  & \begin{tabular}[c]{@{}c@{}}Threshold\\ 0.05\end{tabular} & 89.3 & 15.301/57.828 \\ \hline
					RL-F & $c_{adv}=0.01$ & 90.1 & 51.914/61.215 & $c_{adv}=0.05$ & 84.3 & 34.966/63.480 \\ \hline
				\end{tabular}

	\end{center}

\end{table}

Table \ref{tab:MMM-rst} shows additional result of the 8m and bane\_vs\_bane map. The winning rate of the bane\_vs\_bane map without attack is 84.0\%. For 8m map, OPT outperforms Ra-R/Ra-L and Ru-B, its result is similar to RL-F. For bane\_vs\_bane map, with $\lambda=0.1$, OPT result is similar to Ru-B, and with $\lambda=0.01$ OPT outperforms RL-F.

\begin{table}[h]
	\caption{8m and bane\_vs\_bane map}
	\label{tab:MMM-rst}

	\begin{center}

				\begin{tabular}{c|ccc|ccc}
					\hline
					Map & \multicolumn{3}{c|}{8m Agent 3 and 6} & \multicolumn{3}{c}{bane\_vs\_bane Agent 10} \\ \hline
					\begin{tabular}[c]{@{}c@{}}Attack\\ type\end{tabular} & Parameter & \begin{tabular}[c]{@{}c@{}}Winning\\ rate (\%)\end{tabular} & \begin{tabular}[c]{@{}c@{}}Attacked steps / \\ Total steps\end{tabular} & Parameter & \begin{tabular}[c]{@{}c@{}}Winning\\ rate (\%)\end{tabular} & \begin{tabular}[c]{@{}c@{}}Attacked steps / \\ Total steps\end{tabular} \\ \hline
					\multirow{6}{*}{OPT} & \multirow{3}{*}{$\lambda=1$} & 12.8 & 2.001,1.464/26.848 & \multirow{3}{*}{$\lambda=0.1$} & 80.1 & 7.124/74.455 \\
					& & 8.9 & 2.319,1.459/26.121 & & 79.1 & 9.015/73.818 \\
					& & 8.4 & 2.258,1.467/26.332 & & 81.9 & 7.618/73.672  \\ \cline{2-7} 
					& \multirow{3}{*}{$\lambda=5$} & 21.1 & 1.692,1.428/27.900  & \multirow{3}{*}{$\lambda=0.01$} & 76.0 & 21.749/85.297 \\
					& & 20.3 & 1.675,1.401/27.311  & & 78.5 & 20.925/82.861 \\
					& & 21.4 & 1.721,1.421/27.446  & & 76.7 & 18.996/78.974 \\ \hline
					Ra-R & - & 97.6 & 15\% & - & 81.8 & 15\% \\
					& & & & & 82.8 & 25\% \\ \hline
					Ra-L & - & 88.6 & 15\% & - & 84.1 & 15\% \\
					& & & & & 80.8 & 25\% \\ \hline
					Ru-B & \begin{tabular}[c]{@{}c@{}}Threshold\\ 0.18\end{tabular} & 47.0 & 12.180/72.400  & \begin{tabular}[c]{@{}c@{}}Threshold\\ 0.05\end{tabular} & 79.2 & 9.21/76.463  \\ \hline
					RL-F & $c_{adv}=1$ & 23.0 & 1.896,1.013/28.429 & $c_{adv}=0.01$ & 78.6 & 35.670/83.136 \\ \hline
				\end{tabular}

	\end{center}

\end{table}

Table \ref{tab:25m-rst} shows additional multi agent attack result of 25m and bane\_vs\_bane map. The winning rate of these 2 maps without adversarial attack is 98.5\% and 84.0\% respectively. For 25m map, when attacking 3 agents, OPT can lower the winning rate to around 45\% with each agent less than 10\% of timesteps begin attacked, and it outperforms all baselines including RL-F. For bane\_vs\_bane map, when attacking 3 or 5 agents, OPT can lowerer the winning rate, and it outperforms Ra-R/Ra-L and RL-F. But OPT and Ru-B cannot compare directly, since we cannot find a thshold to let Ru-B attack as many timesteps of agent 0 as OPT does.

\begin{table}
	\caption{25m and bane\_vs\_bane map: attacking multiple agents}
	\label{tab:25m-rst}

	\begin{center}

				\begin{tabular}{c|ccc}
					\hline
					Map & \multicolumn{3}{c}{25m Agent 0, 10, 20} \\ \hline
					\begin{tabular}[c]{@{}c@{}}attack\\ type\end{tabular} & parameter & \begin{tabular}[c]{@{}c@{}}Winning\\ rate (\%)\end{tabular} & \begin{tabular}[c]{@{}c@{}}Attacked steps / \\ Total steps\end{tabular} \\ \hline
					\multirow{3}{*}{OPT} & \multirow{3}{*}{$\lambda=1$} & 43.9 & 4.924,3.491,3.060/51.127 \\
					& & 44.4 & 4.697,3.511,3.269/49.809 \\
					& & 44.8 & 4.826,3.318,2.982/52.125 \\ \hline
					Ra-R & - & 95.4 & 10\%, 10\%, 10\% \\ \hline
					Ra-L & - & 91.4 & 10\%, 10\%, 10\% \\ \hline
					Ru-B & \begin{tabular}[c]{@{}c@{}}Threshold\\ 0.15, 0.1, 0.1\end{tabular} & 56.3 & \begin{tabular}[c]{@{}c@{}}9.622,8.996,6.702\\ /52.410\end{tabular} \\ \hline
					RL-F & $c_{adv}=0.23$ & 80.5 & 10.915,4.781,0.252/51.467 \\ \hline
					Map & \multicolumn{3}{c}{bane\_vs\_bane Agent 0, 10, 20} \\ \hline
					\begin{tabular}[c]{@{}c@{}}Attack\\ type\end{tabular} & Parameter & \begin{tabular}[c]{@{}c@{}}Winning\\ rate (\%)\end{tabular} & \begin{tabular}[c]{@{}c@{}}Attacked steps / \\ Total steps\end{tabular} \\ \hline
					\multirow{3}{*}{OPT} & \multirow{3}{*}{$\lambda=0.01$} & 41.3 & 104.218,21.053,17.397/143.964 \\
					& & 44.9 & 67.105,20.244,22.769/131.688 \\
					& & 27.4 & 102.376,16.529,17.030/160.459 \\ \hline
					Ra-R & - & 90.2 & 75\%, 15\%, 15\% \\ \hline
					Ra-L & - & 91.8 & 75\%, 15\%, 15\% \\ \hline
					Ru-B & \begin{tabular}[c]{@{}c@{}}Threshold\\ 0.005, 0.01, 0.01\end{tabular} & 79.8 & \begin{tabular}[c]{@{}c@{}}54.483,9.809,9.178\\ /78.794\end{tabular} \\ \hline
					RL-F & $c_{adv}=0$ & 71.1 & 63.694,53.639,50.014/92.802 \\ \hline
					Map & \multicolumn{3}{c}{bane\_vs\_bane Agent 0, 5, 10, 15, 20} \\ \hline
					\begin{tabular}[c]{@{}c@{}}Attack\\ type\end{tabular} & Parameter & \begin{tabular}[c]{@{}c@{}}Winning\\ rate (\%)\end{tabular} & \begin{tabular}[c]{@{}c@{}}Attacked steps / \\ Total steps\end{tabular} \\ \hline
					\multirow{3}{*}{OPT} & \multirow{3}{*}{$\lambda=0.01$} & 23.5 & \begin{tabular}[c]{@{}c@{}}119.952,19.067,21.522,15.708,16.029\\ /164.122\end{tabular} \\
					& & 20.2 & \begin{tabular}[c]{@{}c@{}}87.203,25.645,21.332,22.612,22.143\\ /169.967\end{tabular} \\
					& & 26.7 & \begin{tabular}[c]{@{}c@{}}88.507,21.880,16.905,10.793,10.950\\ /163.439\end{tabular} \\ \hline
					Ra-R & - & 90.7 & 70\%, 15\%, 15\%, 15\%, 15\% \\ \hline
					Ra-L & - & 92.2 & 70\%, 15\%, 15\%, 15\%, 15\% \\ \hline
					Ru-B & \begin{tabular}[c]{@{}c@{}}Threshold\\ 0.005,0.005,0.005,0.005,0.005\end{tabular} & 78.5 & \begin{tabular}[c]{@{}c@{}}63.232,9.479,9.048,10.210,11.273\\/84.098\end{tabular} \\ \hline
					RL-F & $c_{adv}=0$ & 67.1 & 37.091,31.063,18.260,27.080,20.191/109.740 \\ \hline
				\end{tabular}
	\end{center}

\end{table}

Table \ref{tab:1c3s5z-vdnqtran-rst} \ref{tab:25m-vdnqtran-rst} shows additional result of attacking VDN and QTRAN policies. During our experiements, we found QTRAN performs not very well in the 25m map, so we only attack VDN policies in the 25m map. The result is similar to QMIX policies. In most cases, OPT outperforms all baselines, and in some cases it outperforms Ru-D.

\begin{table}

	\caption{1c3s5z map VDN / QTRAN}
	\label{tab:1c3s5z-vdnqtran-rst}

	\begin{center}

				\begin{tabular}{c|ccc|ccc}
					\hline
					Map & \multicolumn{3}{c|}{1c3s5z Agent 0 VDN} & \multicolumn{3}{c}{1c3s5z Agent 0 QTRAN} \\ \hline
					\begin{tabular}[c]{@{}c@{}}Attack\\ type\end{tabular} & Parameter & \begin{tabular}[c]{@{}c@{}}Winning\\ rate (\%)\end{tabular} & \begin{tabular}[c]{@{}c@{}}Attacked steps / \\ Total steps\end{tabular} & Parameter & \begin{tabular}[c]{@{}c@{}}Winning\\ rate (\%)\end{tabular} & \begin{tabular}[c]{@{}c@{}}Attacked steps / \\ Total steps\end{tabular} \\ \hline
					\multirow{3}{*}{OPT} & \multirow{3}{*}{$\lambda=5$} & 3.6 & 5.038/60.800  & \multirow{3}{*}{$\lambda=1$} & 0.0 & 4.156/69.583  \\
					& & 4.0 & 5.123/59.584  & & 0.0 & 4.059/67.123  \\
					& & 4.4 & 5.483/60.106  & & 0.0 & 4.005/69.318  \\ \hline
					Ra-R & - & 95.5 & 10\% & & 84.2 & 10\% \\ \hline
					Ra-L & - & 89.4 & 10\% & & 80.0 & 10\% \\ \hline
					Ru-B & \begin{tabular}[c]{@{}c@{}}Threshold\\ 0.025\end{tabular} & 47.4 & 7.652/64.042  & \begin{tabular}[c]{@{}c@{}}Threshold\\ 0.045\end{tabular} & 37.5 & 10.996/86.140 \\ \hline
					RL-F & $c_{adv}=0.3$ & 3.6 & 5.233/60.466 & $c_{adv}=0.1$ & 1.1 & 11.249/69.823 \\ \hline
					Map & \multicolumn{3}{c|}{1c3s5z Agent 1+7 VDN} & \multicolumn{3}{c}{1c3s5z Agent 1+7 QTRAN} \\ \hline
					\begin{tabular}[c]{@{}c@{}}Attack\\ type\end{tabular} & Parameter & \begin{tabular}[c]{@{}c@{}}Winning\\ rate (\%)\end{tabular} & \begin{tabular}[c]{@{}c@{}}Attacked steps / \\ Total steps\end{tabular} & Parameter & \begin{tabular}[c]{@{}c@{}}Winning\\ rate (\%)\end{tabular} & \begin{tabular}[c]{@{}c@{}}Attacked steps / \\ Total steps\end{tabular} \\ \hline
					\multirow{3}{*}{OPT} & \multirow{3}{*}{$\lambda=0.01$} & 5.7 & \begin{tabular}[c]{@{}c@{}}60.854,58.683\\/73.878\end{tabular}  & \multirow{3}{*}{$\lambda=0.01$} & 2.5 & \begin{tabular}[c]{@{}c@{}}59.435,55.070\\/75.038\end{tabular}  \\
					& & 5.0 & \begin{tabular}[c]{@{}c@{}}55.652,57.354\\/75.560\end{tabular}  & & 2.0 & \begin{tabular}[c]{@{}c@{}}59.885,57.792\\/73.920\end{tabular}  \\
					& & 5.4 & \begin{tabular}[c]{@{}c@{}}58.902,58.434\\/72.109\end{tabular} & & 2.3 & \begin{tabular}[c]{@{}c@{}}60.870,58.052,\\/72.261\end{tabular}  \\ \hline
					Ra-R & - & 81.5 & 80\%,80\% & - & 58.8 & 85\%,85\% \\ \hline
					Ra-L & - & 39.4 & 80\%,80\% & - & 17.6 & 85\%,85\% \\ \hline
					Ru-D & - & 24.5 & 100\%,100\% & - & 6.3 & 100\%,100\% \\ \hline
					RL-F & $c_{adv}=0$ & 31.0 & \begin{tabular}[c]{@{}c@{}}79.500,59.505\\/119.208\end{tabular} & $c_{adv}=0$ & 11.9 & \begin{tabular}[c]{@{}c@{}}121.312,44.639\\/138.440\end{tabular} \\ \hline
				\end{tabular}

	\end{center}

\end{table}

\begin{table}
	\caption{25m map VDN}
	\label{tab:25m-vdnqtran-rst}

	\begin{center}

				\begin{tabular}{c|ccc}
					\hline
					Map & \multicolumn{3}{c}{25m Agent 0 VDN} \\ \hline
					\begin{tabular}[c]{@{}c@{}}Attack\\ type\end{tabular} & Parameter & \begin{tabular}[c]{@{}c@{}}Winning\\ rate (\%)\end{tabular} & \begin{tabular}[c]{@{}c@{}}Attacked steps / \\ Total steps\end{tabular} \\ \hline
					\multirow{3}{*}{OPT} & \multirow{3}{*}{$\lambda=0.01$} & 75.6 & 34.392/61.564 \\
					& & 73.1 & 40.409/64.335 \\
					& & 70.2 & 39.407/64.272 \\ \hline
					Ra-R & - & 95.4 & 55\% \\ \hline
					Ra-L & - & 88.4 & 55\% \\ \hline
					Ru-D & - & 81.4 & 100\% \\ \hline
					RL-F & $c_{adv}=0$ & 82.3 & 32.235/55.256 \\ \hline
					Map & \multicolumn{3}{c}{25m Agent 0, 10, 20 VDN} \\ \hline
					\begin{tabular}[c]{@{}c@{}}Attack\\ type\end{tabular} & Parameter & \begin{tabular}[c]{@{}c@{}}Winning\\ rate (\%)\end{tabular} & \begin{tabular}[c]{@{}c@{}}Attacked steps / \\ Total steps\end{tabular} \\ \hline
					\multirow{3}{*}{OPT} & \multirow{3}{*}{$\lambda=1$} & 13.0 & 3.384,3.758,3.199/45.249 \\
					& & 14.3 & 3.749,4.153,2.996/49.728 \\
					& & 11.1 & 3.493,3.648,3.148/45.895 \\ \hline
					Ra-R & - & 96.7 & 10\%, 10\%, 10\% \\ \hline
					Ra-L & - & 93.6 & 70\%, 10\%, 10\% \\ \hline
					Ru-B & \begin{tabular}[c]{@{}c@{}}Threshold\\ 0.02,0.014,0.016\end{tabular} & 82.3 & 4.825,3.816,4.621/42.664 \\ \hline
					RL-F & $c_{adv}=0.2$ & 79.3 & 6.462,6.787,1.411/53.820 \\ \hline
					Map & \multicolumn{3}{c}{25m 0, 5, 10, 15, 20 VDN} \\ \hline
					\begin{tabular}[c]{@{}c@{}}Attack\\ type\end{tabular} & Parameter & \begin{tabular}[c]{@{}c@{}}Winning\\ rate (\%)\end{tabular} & \begin{tabular}[c]{@{}c@{}}Attacked steps / \\ Total steps\end{tabular} \\ \hline
					\multirow{3}{*}{OPT} & \multirow{3}{*}{$\lambda=1$} & 1.3 & \begin{tabular}[c]{@{}c@{}}5.145,4.456,3.901,4.382,3.114\\/48.018\end{tabular} \\
					& & 1.2 & \begin{tabular}[c]{@{}c@{}}4.837,4.328,3.964,4.392,3.177\\/46.169\end{tabular} \\
					& & 0.9 & \begin{tabular}[c]{@{}c@{}}4.835,4.805,4.206,4.103,3.240\\/49.859\end{tabular} \\ \hline
					Ra-R & - & 95.6 & 10\%, 10\%, 10\%, 10\%, 10\% \\ \hline
					Ra-L & - & 86.7 & 10\%, 10\%, 10\%, 10\%, 10\% \\ \hline
					Ru-B & \begin{tabular}[c]{@{}c@{}}Threshold\\ 0.019,0.015,0.015,0.015,0.015\end{tabular} & 50 & \begin{tabular}[c]{@{}c@{}}9.677,5.401,6.301,5.445,5.343\\/54.064\end{tabular} \\ \hline
					RL-F & $c_{adv}=0.1$ & 9.3 & 5.370,12.808,18.994,11.546,30.295/116.817 \\ \hline
				\end{tabular}
	\end{center}

\end{table}

\end{document}